\documentclass{article}

\PassOptionsToPackage{numbers, compress}{natbib}

\usepackage{alex404}
\usepackage[preprint]{neurips_2019}

\usepackage[utf8]{inputenc} 
\usepackage[T1]{fontenc}    
\usepackage{hyperref}       
\usepackage{url}            
\usepackage{booktabs}       
\usepackage{amsfonts}       
\usepackage{nicefrac}       
\usepackage{microtype}      
\usepackage{amsmath,amsthm,mathtools,thmtools,caption,xcolor}
\usepackage[skip=0pt]{subcaption}

\title{Conditional Finite Mixtures of Poisson Distributions for Context-Dependent Neural Correlations}

\author{%
  Sacha Sokoloski \\
  Systems and Computational Biology\\
  Albert Einstein College of Medicine\\
  The Bronx, NY 10461 \\
  \texttt{sacha.sokoloski@einstein.yu.edu} \\
  \And{}
  Ruben Coen-Cagli \\
  Systems and Computational Biology\\
  Albert Einstein College of Medicine\\
  The Bronx, NY 10461 \\
  \texttt{ruben.coen-cagli@einstein.yu.edu} \\
}

\theoremstyle{plain}
\newtheorem{theorem}{Theorem}

\newcommand{\inv}{^{-1}}

\hypersetup{
    colorlinks=true,
    menucolor=black,
    citecolor=blue,
    linkcolor=red
}

\begin{document}

\maketitle

\begin{abstract}
    Parallel recordings of neural spike counts have revealed the existence of context-dependent noise correlations in neural populations. Theories of population coding have also shown that such correlations can impact the information encoded by neural populations about external stimuli. Although studies have shown that these correlations often have a low-dimensional structure, it has proven difficult to capture this structure in a model that is compatible with theories of rate coding in correlated populations. To address this difficulty we develop a novel model based on conditional finite mixtures of independent Poisson distributions. The model can be conditioned on context variables (e.g.\ stimuli or task variables), and the number of mixture components in the model can be cross-validated to estimate the dimensionality of the target correlations. We derive an expectation-maximization algorithm to efficiently fit the model to realistic amounts of data from large neural populations. We then demonstrate that the model successfully captures stimulus-dependent correlations in the responses of macaque V1 neurons to oriented gratings. Our model incorporates arbitrary nonlinear context-dependence, and can thus be applied to improve predictions of neural activity based on deep neural networks.
\end{abstract}

\section{Introduction}

Computational neuroscientists have made significant progress in understanding how correlations amongst neural spike-counts influence information processing~\cite{averbeck_neural_2006,kohn_correlations_2016}. In particular, correlated fluctuations in the responses to a fixed stimulus (termed noise correlations) depend on stimulus identity, and can both hinder and facilitate information processing in model neural circuits~\cite{abbott_effect_1999,sompolinsky_population_2001,shamir_implications_2006,ecker_effect_2011,moreno-bote_information-limiting_2014}. However, due to the challenges in assessing the strength and order of noise correlations empirically, it remains unclear how noise correlations affect computation in biological neural circuits.

Measuring correlations of every order in large neural populations is infeasible, and modelling neural correlations requires knowledge of the order and dimensionality of significant correlations. Most models measure exclusively pair-wise correlations~\cite{schneidman_towards_2016,gardella_modeling_2018}, which can often explain most of the variability in spike-count data~\cite{schneidman_weak_2006,granot-atedgi_stimulus-dependent_2013}. Moreover, dimensionality-reduction methods have shown that the complete set of pair-wise neural correlations often has a low-dimensional structure for both total~\cite{cowley_stimulus-driven_2016} and noise~\cite{ecker_state_2014,cunningham_dimensionality_2014,goris_partitioning_2014,okun_diverse_2015,rosenbaum_spatial_2017} correlations. Nevertheless, higher-order correlations become more significant as both stimulus complexity and neural population size increase~\cite{schneidman_towards_2016}, motivating the use of higher-order models when modelling noise correlations in large populations of neurons.

Dimensionality reduction methods have provided many insights into neural correlations, but correlations must be studied through models of population codes in order to understand their effect on neural coding. Maximum entropy models of binary neural spiking drive much of the contemporary research on neural correlations in model population codes~\cite{schneidman_towards_2016}. The flexibility of the maximum entropy framework allows the techniques for analyzing and fitting pair-wise models to be generalized to both sparse, higher-order models~\cite{ganmor_sparse_2011} and stimulus-dependent models~\cite{granot-atedgi_stimulus-dependent_2013}. Nevertheless, there is still a need for a rate-based model of population coding that can reliably capture the complex yet low-dimensional noise correlations found in neural data.

To address this we propose a conditional maximum entropy model, specifically a conditional finite mixture of independent Poisson distributions (CMP). When conditioned on a context variable (e.g.\ a stimulus) a CMP is a mixture of component collections of independent Poisson distributions~\cite{karlis_finite_2007}. A CMP with only one component reduces to a standard rate-based population code with Poisson neurons that are conditionally independent given the stimulus~\cite{ma_bayesian_2006,ostojic_spiking_2011}, and adding components to the CMP introduces noise correlations with dimensionality controlled by the number of components.

By combining the properties of maximum entropy models with additional features of Poisson neurons we derive a hybrid expectation-maximization algorithm that can efficiently and reliably fit the CMP to data. We apply the CMP to neural population data recorded in macaque primary visual cortex (V1). We find that noise correlation structure is best captured with 3--5 components across several datasets, and that while overall dimensionality of correlations is largely stimulus-independent, the structure of noise correlations (i.e.\ the relative weights of the mixture components) depends on the stimulus.



\section{Conditional Finite Mixtures of Independent Poisson Distributions}\label{sec:cmps}

Many of the equations that describe how to analyze and train the CMP model can be solved by expressing the parameters of the model in an appropriate coordinate system. Two of the coordinate systems we consider are the mean and natural parameters that arise in the context of exponential families\footnote{Maximum entropy models are also known as exponential families, and we prefer that name in this paper.}. We thus begin this section by reviewing exponential families, primarily for the purpose of developing notation (for a thorough development see~\cite{amari_methods_2007,wainwright_graphical_2008}). We continue by showing that the standard weighted-sum form of a finite mixture model is a third parameterization of a kind of exponential family known as an exponential family harmonium~\cite{welling_exponential_2005}. Finally, we introduce and develop conditional finite mixture models and CMPs, and we derive training algorithms for such models including the hybrid expectation-maximization algorithm for training CMPs.

\subsection{Exponential Families}

Consider a random variable $X \in \mathcal X$ with an unknown distribution $P_X$, and suppose we are given an independent and identically distributed sample ${\{X_i\}}_{i=1}^{m_S}$ from $P_X$ such that every $X_i \sim P_X$. We may model $P_X$ based on the sample ${\{X_i\}}_{i=1}^{m_S}$ by first defining a statistic $\V s_X \colon \mathcal X \to \Eta_X$, where $\Eta_X$ is the codomain of $\V s_X$ with dimension $m_X$. We then look for a probability distribution $Q_X$ that satisfies $\E_Q[\V s_X(X)] = \frac{1}{m_S}\sum_{i=1}^{m_S} \V s_X(X_i)$, where $\E_Q[f(X)] = \int_{\mathcal X} f dQ_X$ is the expected value of $f(X)$ under $Q_X$. This is of course an under-constrained problem, but if we also assume that $Q_X$ must have maximum entropy, then we arrive at a family of distributions which uniquely satisfy the constraints for every value of $\frac{1}{m_S}\sum_{i=1}^{m_S} \V s_X(X_i)$.

An  $m_X$-dimensional \emph{exponential family} $\mathcal M_X$ is defined by the \emph{sufficient statistic} $\V s_X$, as well as a \emph{base measure} $\mu_X$ which helps define integrals and expectations within the family. An exponential family is parameterized by a set of \emph{natural parameters} $\Theta_X$, such that each element of the family $Q_X \in \mathcal M_X$ may be identified with some parameters $\eprms_X \in \Theta_X$. The density $q_X$ of the distribution $Q_X$ is given by $\log q_X(x) = \V s_X(x) \cdot \eprms_X - \psi_X(\eprms_X)$, where $\psi_X(\eprms_X) = \log \int_{\mathcal X}e^{\V s_X(x) \cdot \eprms_X}\mu_X(dx)$ is the log-partition function. Expectations of any $Q_X \in \mathcal M_X$ are then given by $\E_Q[f(X)] = \int_{\mathcal X} f dQ_X = \int_{\mathcal X} f \cdot q_X d\mu_X$. Because each $Q_X \in \mathcal M_X$ is uniquely defined by $\E_Q[\V s_X(X)]$, the means of the sufficient statistic also parameterize $\mathcal M_X$. The space of all \emph{mean parameters} is $\Eta_X$, and we denote them by $\mprms_X = \E_Q[\V s_X(X)]$. Finally, a sufficient statistic is \emph{minimal} when its component functions are non-constant and linearly independent. If the sufficient statistic $\V s_X$ of a given family $\mathcal M_X$ is minimal, then $\Theta_X$ and $\Eta_X$ are isomorphic. Moreover, the transition functions between them $\V \tau_X \colon \Theta_X \to \Eta_X$ and $\V \tau_X\inv \colon \Eta_X \to \Theta_X$ are given by $\V \tau_X(\eprms_X) = \partial_{\eprms_X}\psi_X(\eprms_X)$, and $\V \tau_X\inv(\mprms_X) = \partial_{\mprms_X} \phi_X(\mprms_X)$, where $\phi_X(\mprms_X) = \E_Q[\log q_X(X)]$ is the negative entropy of $Q_X$.

Before we conclude this subsection, we highlight the exponential families of \emph{categorical distributions} $\mathcal M_C$ and of \emph{independent Poisson distributions} $\mathcal M_N$. The $m_C$-dimensional categorical exponential family $\mathcal M_C$ contains all the distributions over integer values between 0 and $m_C$. The base measure of $\mathcal M_C$ is the counting measure, and the $j$th element of its sufficient statistic $\V s_C(j)$ at $j$ is 1, and 0 for any other elements. The sufficient statistic $\V s_C$ is thus a vector of all zeroes when $j=0$, and all zeroes except for the $j$th element when $j > 0$. The $m_N$-dimensional family of independent Poisson distributions $\mathcal M_N$ contains distributions over $m_N$-dimensional vectors of natural numbers, where each element of the vector is independently Poisson distributed. The sufficient statistic of $\mathcal M_N$ is the identity function, and the base measure is given by $\mu_N(\{\V n\}) = \prod_{j=1}^{m_N}\frac{1}{n_j!}$.

\subsection{Finite Mixture Models}

In this subsection we show how finite mixture models can be expressed as latent-variable models such that the complete joint model is a particular kind of exponential family. To begin, let us consider an exponential family $\mathcal M_X$. The density of a \emph{finite mixture distribution} $Q_X^*$ of $m_C + 1$ distributions in $\mathcal M_X$ has the form $q_X^*(x) = \sum_{j=0}^{m_C} \eta^*_{C,j} q^*_{X,j}(x)$, where each distribution $Q^*_{X,j} \in \mathcal M_X$ is a \emph{mixture component} with natural parameters $\eprms^*_{X,j}$, each $\eta^*_{C,j}$ is a \emph{mixture weight}, and the weights satisfy the constraints $0 < \eta^*_{C,j} < 1$ and $\eta^*_0 = 1 - \sum_{j=1}^{m_C} \eta^*_{C,j}$.

Building a model out of mixture distributions is theoretically problematic because swapping component indices has no effect on a mixture distribution, and their parameters are therefore not identifiable. Nevertheless, we may express mixture distributions as latent-variable distributions with the identifiable form $\sum_{j=0}^{m_C} \eta^*_{C,j} q^*_{X,j}(x) = \sum_{j=0}^{m_C} q^*_C(j) q^*_{X \mid C}(x \mid j) = \sum_{j=0}^{m_C} q^*_{XC}(x,j)$, where $\eta^*_{C,j} = q^*_C(j)$ and $q^*_{X,j}(x) = q^*_{X \mid C}(x \mid j)$. Moreover, $Q_C^*$ is a categorical distribution with mean parameters $\mprms_C^* = {(\eta^*_{C,j})}_{j=1}^{m_C}$, and $Q_C^*$ may thus be described by the natural parameters $\eprms_C^* = \V \tau_C\inv(\mprms_C^*)$. Given $\mathcal M_X$ and $m_C$, we therefore define a \emph{finite mixture model} $\mathcal M_{XC}^*$ as the set of all distributions $Q^*_{XC}$, and parameterize it by the mixture weights $\eprms_C^*$ and mixture components parameters ${\{\eprms^*_{X,j}\}}_{j=0}^{m_C}$.

Now, an \emph{exponential family harmonium} is a kind of product exponential family which includes restricted Boltzmann machines as a special case~\cite{welling_exponential_2005}. We may construct an exponential family harmonium $\mathcal M_{XY}$ out of $\mathcal M_X$ and $\mathcal M_Y$ by defining the base measure of $\mathcal M_{XY}$ as the product measure $\mu_X \times \mu_Y$, and by defining the sufficient statistic of $\mathcal M_{XY}$ as the vector which contains the concatenation of all the elements in $\V s_X$, $\V s_Y$, and the outer product $\V s_X \otimes \V s_Y$. Intuitively, $\mathcal M_{XY}$ contains all the distributions with densities of the form $q_{XY}(x,y) \propto e^{\V s_X(x) \cdot \eprms_X + \V s_Y(y) \cdot \eprms_Y + \V s_X(x) \cdot \iprms_{XY} \cdot \V s_Y(y)}$, where $\eprms_X$, $\eprms_Y$, and $\iprms_{XY}$ are the natural parameters of $Q_{XY} \in \mathcal M_{XY}$. As we show, a harmonium defined partially by the categorical exponential family is equivalent to a mixture model.

\begin{theorem}\label{thm:mixture}
    Let $\mathcal M_X$ be an exponential family, let $\mathcal M_C$ be the categorical exponential family of dimension $m_C$, let $\mathcal M_{XC}^*$ be the mixture model defined by $\mathcal M_X$ and $m_C$, and let $\mathcal M_{XC}$ be the exponential family harmonium defined by $\mathcal M_X$ and $\mathcal M_C$. Then $Q_{XC} \in \mathcal M_{XC}$ with natural parameters $\eprms_X$, $\eprms_C$, $\iprms_{XC}$ and $Q_{XC} \in \mathcal M_{XC}^*$ with mixture parameters $\eprms^*_C$ and ${\{\eprms^*_{X,j}\}}_{j=0}^{m_C}$ if and only if
        \begin{align}
        \theta^*_{C,j} &= \theta_{C,j} + \psi_X(\eprms_{X,j} + \eprms_X) - \psi_X(\eprms_X),\label{eq:components-1} \\
        \eprms^*_{X,j} &= \eprms_{X,j} + \eprms_X,\label{eq:components-2}
    \end{align}
    where $\theta_{C,j}$ and $\theta^*_{C,j}$ are the $j$th elements of $\eprms_C$ and $\eprms_C^*$, $\eprms_{X,j}$ is the $j$th row of $\iprms_{XC}$, and $\eprms_{X,0} = \V 0$.
    \begin{proof}
        We identify $Q_{XC}$ and $Q^*_{XC}$ by identifying the parameters of their conditionals and marginals. First note that the conditional distribution $Q_{X \mid C = j}$ at $j$ is an element of $\mathcal M_X$ with parameters $\eprms_X + \iprms_{XC} \cdot \V s_C(j)$~\cite{welling_exponential_2005}, and that the sufficient statistic $\V s_C$ in this expression essentially selects a row from $\iprms_{XC}$, or returns $\V 0$. Thus, where $\eprms_{X,j}$ is the $j$th row of $\iprms_{XC}$, $Q_{X \mid C = j} = Q^*_{X \mid C = j}$ if and only if $\eprms^*_{X,0} = \eprms_X$, and $\eprms^*_{X,j} = \eprms_{X,j} + \eprms_X$ for $j > 0$.

        To equate $Q^*_C$ and $Q_C$, first note that the marginal density of a harmonium distribution satisfies $\log q_C(j) = \eprms_C \cdot \V s_C(j) + \psi_X(\eprms_X + \iprms_{XC} \cdot \V s_C(j)) - \psi_{XC}(\eprms_X, \eprms_C, \iprms_{XC})$~\cite{welling_exponential_2005}. Because $\mathcal M_{XC}$ is partially defined by $\mathcal M_C$, we may show with a bit of algebra that $\psi_{XC}(\eprms_X, \eprms_C, \iprms_{XC}) = \psi_X(\eprms_X) + \psi_C(\eprms_C^*)$, where $\eprms^*_C$ satisfies Equation~\ref{eq:components-1}, and by substituting this into the expression for the harmonium marginal, we may conclude that $Q_C = Q^*_C$ if and only if Equation~\ref{eq:components-1} is satisfied.
    \end{proof}
\end{theorem}

The advantage of formulating finite mixture models as exponential family harmoniums is that we may apply the theory of harmoniums and exponential families to analyzing and training them. For example, given a sample ${\{X_i\}}_{i=1}^{m_S}$ and a harmonium distribution $Q_{XY} \in \mathcal M_{XY}$ with parameters $\eprms_X$, $\eprms_Y$, and $\iprms_{XY}$, an iteration of the expectation-maximization algorithm (EM) may be formulated as:
\begin{description}
    \item[Expectation Step:] compute the unobserved means $\mprms_{Y,i} = \V \tau_Y(\eprms_Y + \V s_X(X_i) \cdot \iprms_{XY})$ for every $i$,
    \item[Maximization Step:] evaluate $\V \tau\inv_{XY}(\frac{1}{m_S} \sum_{i=1}^{m_S} \V s_X(X_i), \frac{1}{m_S} \sum_{i=1}^{m_S} \mprms_{Y,i}, \frac{1}{m_S} \sum_{i=1}^{m_S} \V s_X(X_i) \otimes \mprms_{Y,i})$.
\end{description}
On the other hand, the stochastic log-likelihood gradients of the parameters of $Q_{XY}$ are
\begin{align}
    \partial_{\eprms_X} \log q_X(X_i) &= \V s_X(X_i) - \mprms_X, \label{eq:harmonium-gradient-1} \\
    \partial_{\eprms_Y} \log q_X(X_i) &= \V \tau_Y(\eprms_Y + \V s_X(X_i) \cdot \iprms_{XY}) - \mprms_Y, \label{eq:harmonium-gradient-2} \\
    \partial_{\iprms_{XY}} \log q_X(X_i) &= \V s_X(X_i) \otimes \V \tau_Y(\eprms_Y + \V s_X(X_i) \cdot \iprms_{XY}) - \V H_{XY}. \label{eq:harmonium-gradient-3}
\end{align}
where $(\mprms_X, \mprms_Y, \V H_{XY}) = \V \tau_{XY}(\eprms_X, \eprms_Y, \iprms_{XY})$.

\begin{figure}
\begin{minipage}[c]{0.4\textwidth}
    \centering
    \includegraphics[width=\textwidth]{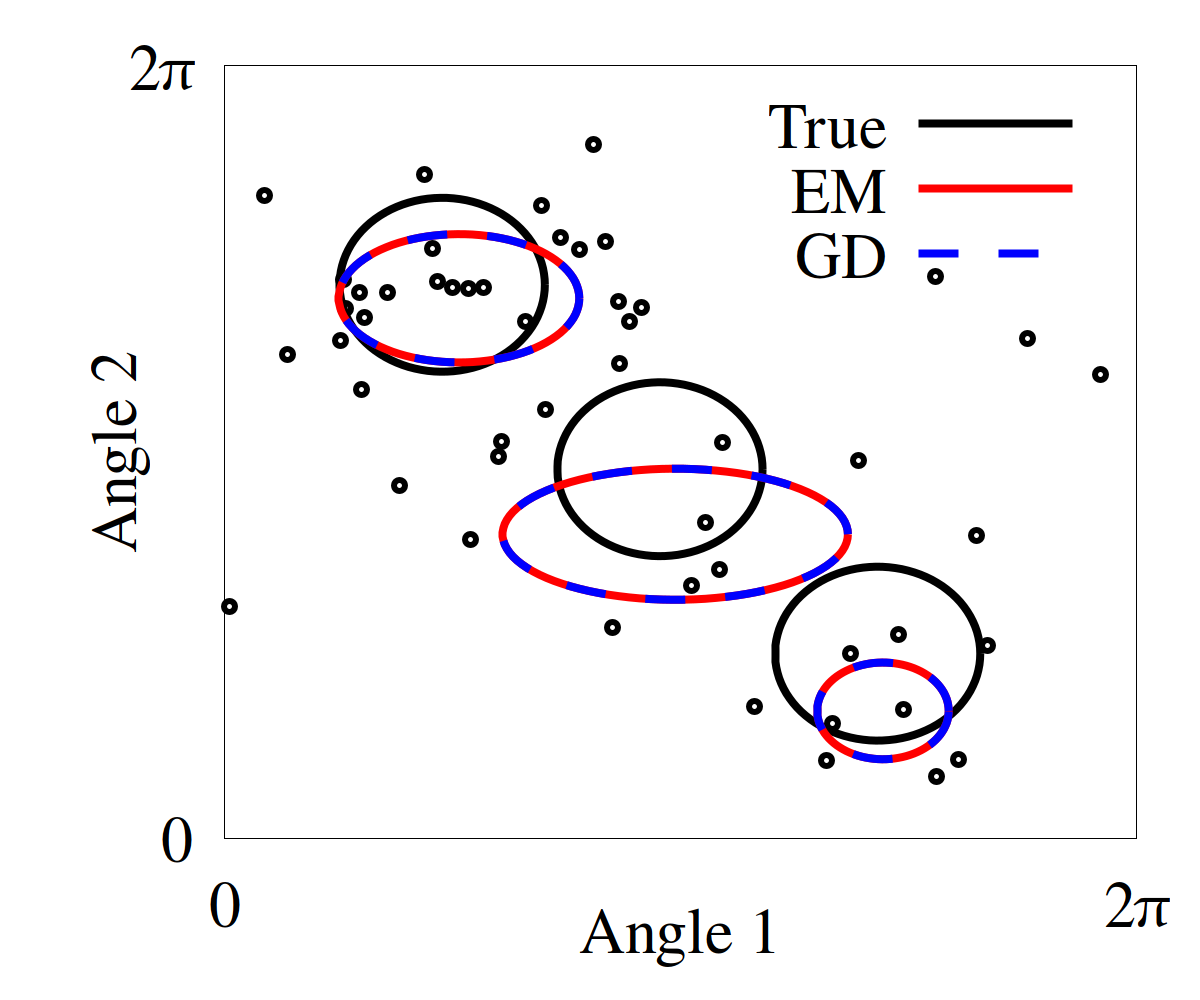}
\end{minipage}
\hfill
\begin{minipage}[c]{0.54\textwidth}
    \centering
    \captionof{figure}{Training a finite mixture of an independent product of 2 von Mises distributions on synthetic data with EM vs batch gradient descent (GD). Ellipsoids indicate the precisions of the component densities of the true mixture (black), the EM-trained mixture (red), and the GD-trained mixture (dashed blue). Black circles are synthetic data. EM and GD find the same solutions starting from the same initial conditions and with similar computation time.}\label{fig:mixture-model}
\end{minipage}
\end{figure}

Whether or not the various transition functions and their inverses can be evaluated depends on the defining manifolds $\mathcal M_X$ and $\mathcal M_Y$. When $\mathcal M_Y = \mathcal M_C$ is the categorical family, its transition function $\V \tau_C$ is computable, and whether $\V \tau_{XY}$ or $\V \tau_{XY}\inv$ are computable is reducible to whether $\V \tau_X$ or $\V \tau_Y\inv$ are computable, respectively. EM is typically preferred when maximizing the likelihood of finite mixture model parameters, especially for finite mixtures of normal distributions. However, gradient descent can perform just as well as EM for certain mixture models, as we demonstrate in figure~\ref{fig:mixture-model}.

\subsection{Conditional Finite Mixture Models}\label{sec:conditional-mm}

According to Equations~\ref{eq:components-1} and~\ref{eq:components-2} we may induce context dependence in all the mixture parameters of a mixture distribution $Q_{XC} \in \mathcal M_{XC}$ by allowing $\eprms_X$ to depend on additional variables. We thus define a \emph{conditional finite mixture model} $\mathcal M_{XC \mid Z}$ as the set of all the conditional distributions $Q_{XC \mid Z}$ with densities $q_{XC \mid Z}(x,j \mid z)\propto e^{\V s_X(x) \cdot (\eprms_X + \eprms_{X \mid Z}(z)) + \eprms_C \cdot \V s_C(j) + \V s_X(x) \cdot \iprms_{XC} \cdot \V s_C(j)}$, where $\eprms_{X \mid Z} \colon \mathcal Z \to \Theta_X$ is defined by an additional set of parameters $\rprms$. In the conditional setting we aim to maximize the conditional log-likelihood $\sum _{i=1}^{m_S} q_{X \mid Z}(X_i \mid Z_i)$ of the parameters $\eprms_X$, $\eprms_C$, $\iprms_{XC}$, and $\rprms$, given a sample ${\{(X_i,Z_i)\}}_{i=1}^{m_S}$ from a target conditional distribution $P_{X \mid Z}$. Stochastic gradient ascent on the conditional log-likelihood remains relatively straightforward; given a sample point $(X_i, Z_i)$, one must simply backpropagate the error-gradient $S_X(X_i) - \mprms_X(Z_i)$ through the function $\eprms_{X \mid Z}$, where $(\mprms_X(Z_i), \mprms_C(Z_i), \V H_{XC}(Z_i)) = \V \tau_{XC}(\eprms_X + \eprms_{X \mid Z}(Z_i),\eprms_C, \iprms_{XC})$.

The expectation step of the EM algorithm also remains trivial for conditional mixtures. It is easy to show that for any $Q_{XC \mid Z} \in \mathcal M_{XC \mid Z}$ the conditional distribution $Q_{C \mid X, Z} = Q_{C \mid X}$, and we may therefore continue to evaluate the unobserved means as $\mprms_{C,i} = \V \tau_C(\eprms_C + \V s_X(X_i) \cdot \iprms_{XC})$. On the other hand, the maximization step can be expressed as maximizing the sum over $i$ of the function
\begin{multline}\label{eq:maximization-step}
    \mathcal L(\eprms_X, \eprms_C, \iprms_{XC}, \rprms, \mprms_{C,i}, X_i, Z_i) = \V s_X(X_i) \cdot (\eprms_X + \eprms_{X \mid Z}(Z_i; \rprms)) \\ + \eprms_C \cdot \mprms_{C,i} + \V s_X(X_i) \cdot \iprms_{XC} \cdot \mprms_{C,i} - \psi_{XC}(\eprms_X + \eprms_{X \mid Z}(Z_i; \rprms), \eprms_C, \iprms_{XC})
\end{multline}
with respect to the parameters $\eprms_X$, $\eprms_C$, $\iprms_{XC}$, and $\rprms$. It is more difficult to evaluate the maximization step for conditional finite mixtures due to the dependence on $Z_i$, although we may still find maxima of the objective by gradient ascent. Nevertheless, when $\mathcal M_X = \mathcal M_N$ is the family of $m_N$ independent Poisson distributions, we can fact evaluate the maximization step in closed-form with respect to the parameters $\eprms_N$ and $\iprms_{NC}$. We refer to the conditional finite mixture model $\mathcal M_{NC \mid Z}$ as the family of \emph{conditional finite mixtures of independent Poisson distributions} (CMPs).

\begin{theorem}\label{thm:partial-update}
    Let $\mathcal M_{NC \mid Z}$ be a CMP model defined by $m_C$, $m_N$, and $\mathcal Z$. Then
    \begin{equation*}
        \argmax_{\eprms_N} \sum_{i=1}^{m_S}\mathcal L(\eprms_N, \eprms_C, \iprms_{NC}, \rprms, \mprms_{C,i}, N_i, Z_i) = \eprms^\dagger_{N,0},
    \end{equation*}
    and
    \begin{equation*}
        \argmax_{\eprms_{N,j}} \sum_{i=1}^{m_S}\mathcal L(\eprms_N, \eprms_C, \iprms_{NC}, \rprms, \mprms_{C,i}, N_i, Z_i) = \eprms^\dagger_{N,j} - \eprms^\dagger_{N,0},
    \end{equation*}
    where
    \begin{equation*}
        \theta^\dagger_{N,j,k} = \log \Big(\frac{\sum_{i=1}^{m_S} \eta_{C,i,j} N_{i,k}}{\sum_{i=1}^{m_S} \eta_{C,i,j} e^{\theta_{N \mid Z,k}(Z_i)}}\Big).
    \end{equation*}
    \begin{proof}
        Similar to Theorem~\ref{thm:mixture}, consider the change of variables $\eprms^\dagger_{N,0} = \eprms_N$,$\eprms^\dagger_{N,j} = \eprms_{N,j} + \eprms_N$, and $\theta^\dagger_{C,j}(Z_i) = \theta_{C,j} + \psi_N(\eprms_{N,j} + \eprms_N + \eprms_{N \mid Z}(Z_i)) - \psi_N(\eprms_N + \eprms_{N \mid Z}(Z_i))$. We may express our optimization in this alternative parameterization as
\begin{multline*}
    \mathcal L_i^\dagger({\{\eprms^\dagger_{N,j}\}}_{j=0}^{m_C}) = \sum_{j=0}^{m_C} \eta_{C,i,j} \big(N_i \cdot \eprms^\dagger_{N,j} - \psi_N(\eprms^\dagger_{N,j} + \eprms_{N \mid Z}(Z_i))\big)\\ + N_i \cdot \eprms_{N \mid Z}(Z_i) + \eprms^\dagger_C(Z_i) \cdot \mprms_{C,i} - \psi_C(\eprms^\dagger_C(Z_i)),
\end{multline*}
such that $\max_{\eprms_N, \iprms_{NC}} \sum_{i=1}^{m_S}\mathcal L = \max_{{\{\eprms^\dagger_{N,j}\}}_{j=0}^{m_C}} \sum_{i=1}^{m_S} \mathcal L_i^\dagger$. Moreover, the log-partition function of $\mathcal M_N$ is given by $\psi_N(\eprms_N) = \sum_{k=1}^{m_N} e^{\theta_{N,k}}$ and we may therefore conclude that
        \begin{align*}
            \argmax_{\theta^\dagger_{N,j,k}} \sum_{i=1}^{m_S}\mathcal L_i^\dagger({\{\eprms^\dagger_{N,j}\}}_{j=0}^{m_C}) &= \argmax_{\theta^\dagger_{N,j,k}} \Big \{ \sum_{i=1}^{m_S} \eta_{C,i,j} \big(N_{i,k} \theta^\dagger_{N,j,k} - e^{ \theta^\dagger_{N,j,k} + \theta_{N \mid Z,k}(Z_i)}\big) \Big \} \\
                                                                                                                          &= \argmax_{\theta^\dagger_{N,j,k}} \Big \{ \theta^\dagger_{N,j,k} \frac{\sum_{i=1}^{m_S} \eta_{C,i,j} N_{i,k}}{\sum_{i=1}^{m_S} \eta_{C,i,j} e^{\theta_{N \mid Z, k}(Z_i)}} - e^{ \theta^\dagger_{N,j,k}} \Big \}.
        \end{align*}
        This is an $\argmax$ variant of a Legendre transform and its solution is $\log \big( \frac{\sum_{i=1}^{m_S} \eta_{C,i,j} N_{i,k}}{\sum_{i=1}^{m_S} \eta_{C,i,j} e^{\theta_{N \mid Z, k}(Z_i)}}\big)$.
    \end{proof}
\end{theorem}

\subsection{Training Algorithms}\label{sec:training-algorithms}

Based on these results, we propose three algorithms for training conditional finite mixture models, the third of which is specific to CMPs, and we abbreviate them as EM, SGD, and Hybrid. The first approximates expectation-maximization (EM) by computing the expectation step in closed-form, and approximating the maximization step by gradient ascent of the objective in Definition~\ref{eq:maximization-step}. The second is stochastic gradient descent (SGD) of the negative log-likelihood based on Equations~\ref{eq:harmonium-gradient-1},~\ref{eq:harmonium-gradient-2}, and~\ref{eq:harmonium-gradient-3}. The third is the Hybrid algorithm based on Theorem~\ref{thm:partial-update}. The Hybrid algorithm alternates between one of two phases every training epoch: the first phase is simply SGD with respect to all the CMP parameters, and the second phase is to update the parameters $\eprms_N$ and $\iprms_{NC}$ based on Theorem~\ref{thm:partial-update}.

\section{Applications to Synthetic and Real Data}\label{sec:applications}

\begin{figure}
\begin{minipage}[c]{0.38\textwidth}
    \centering
    \includegraphics[width=\textwidth]{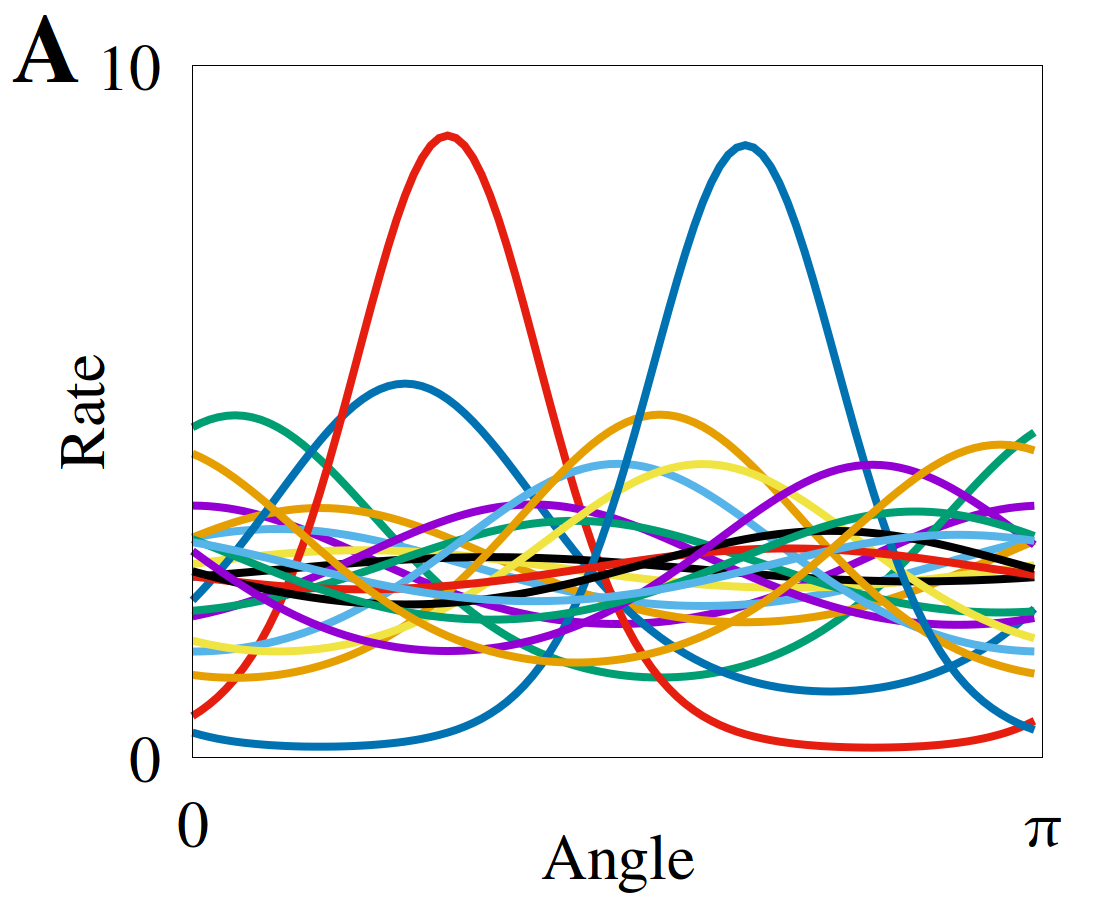}
    \includegraphics[width=\textwidth]{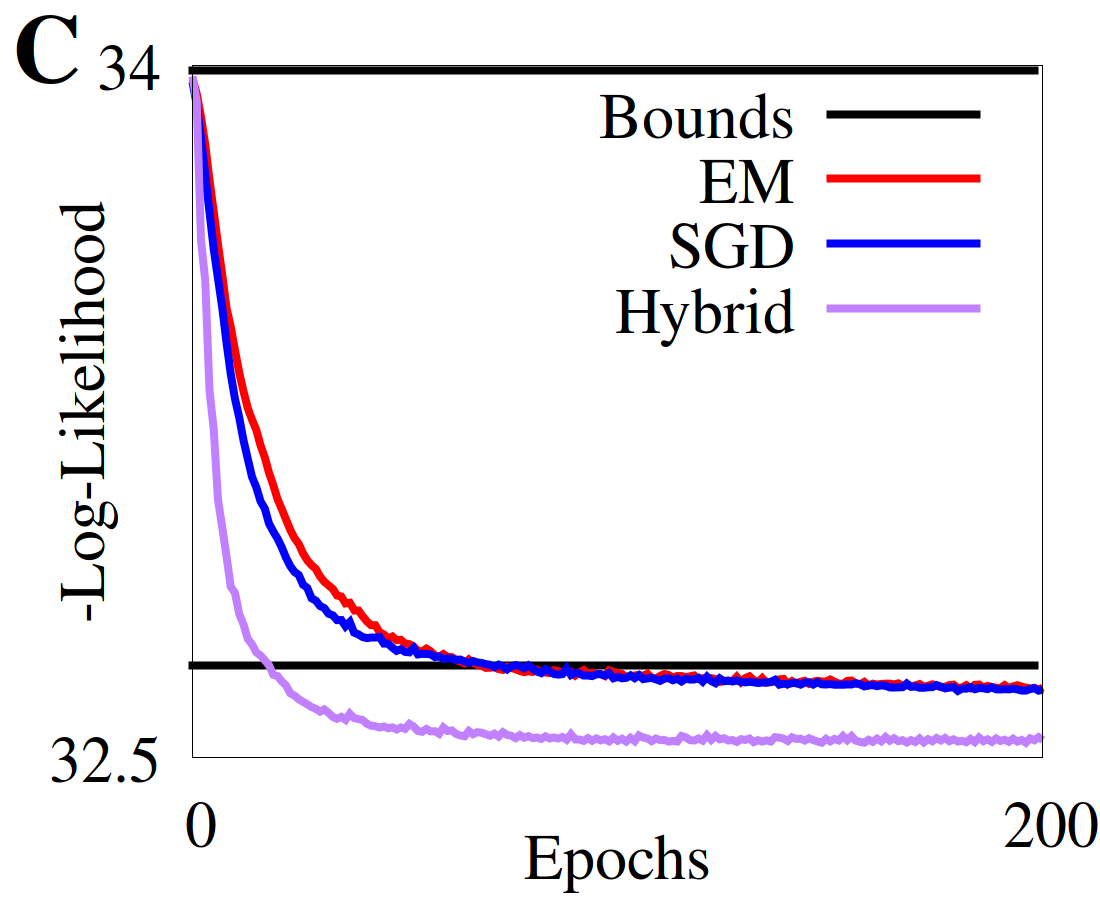}
\end{minipage}
\begin{minipage}[c]{0.38\textwidth}
    \centering
    \includegraphics[width=\textwidth]{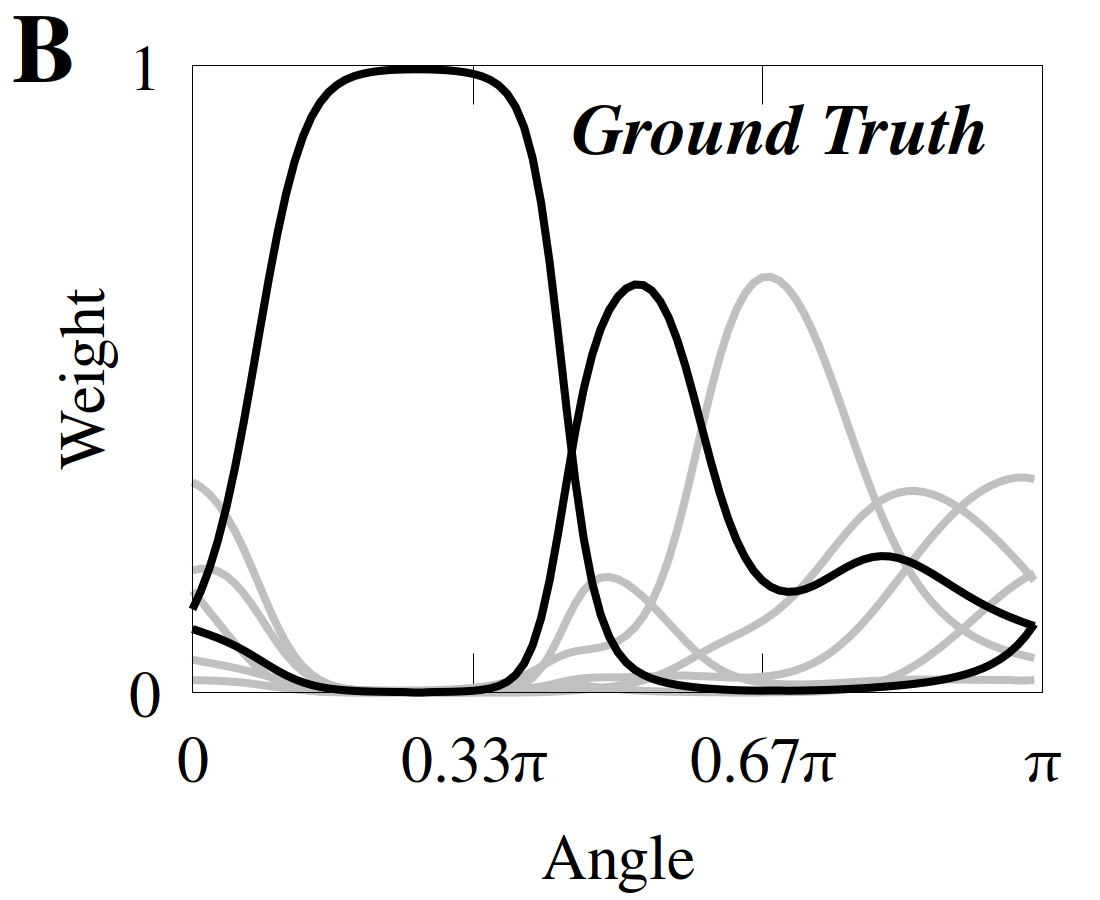}
    \includegraphics[width=\textwidth]{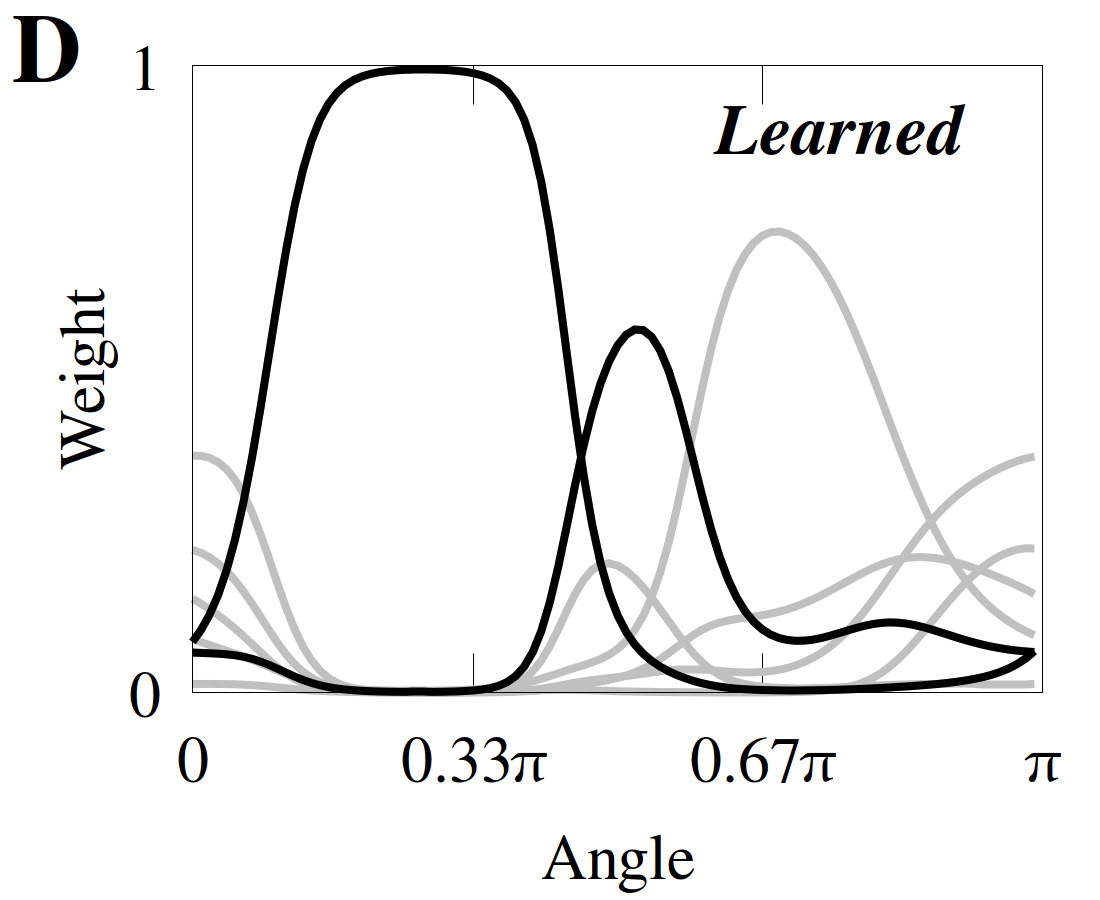}
\end{minipage}
\hfill
\begin{minipage}[c]{0.22\textwidth}
    \includegraphics[width=\textwidth]{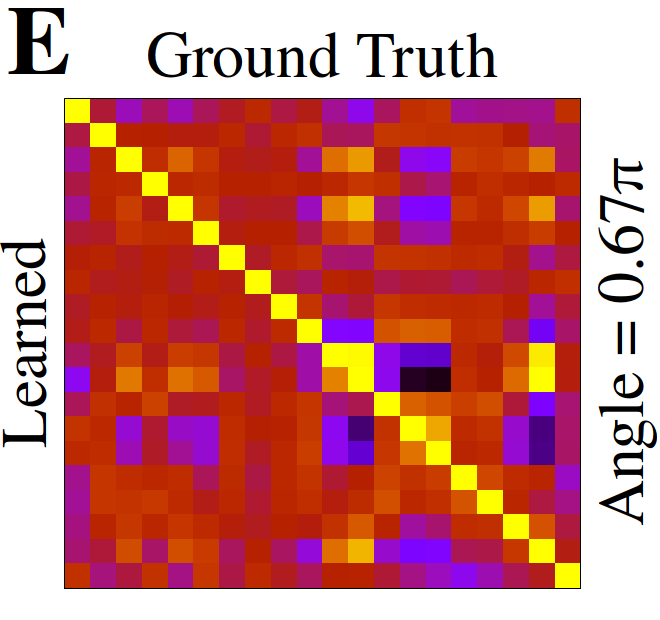}
    \includegraphics[width=\textwidth]{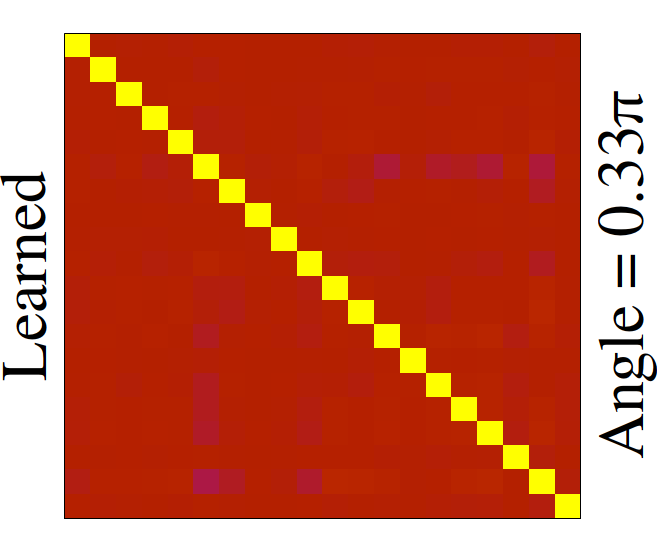}
    \includegraphics[width=\textwidth]{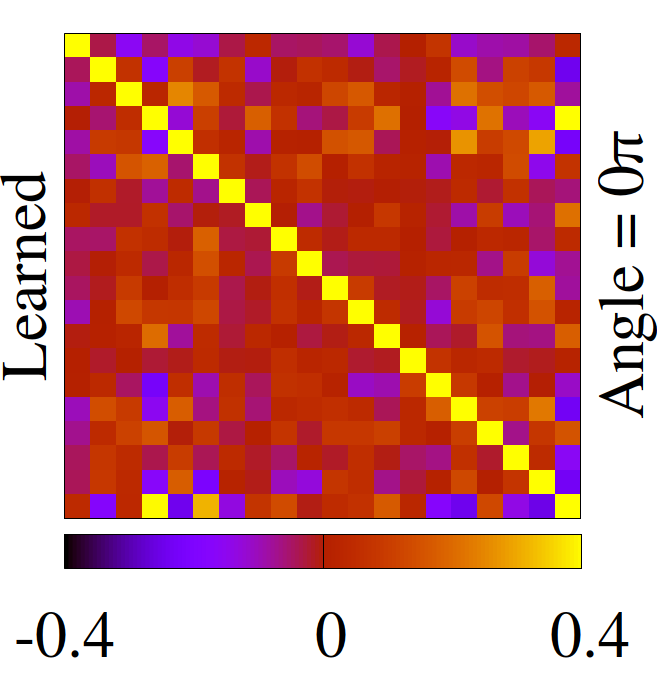}
\end{minipage}

\caption{Training a CMP model on synthetic data generated from a ground truth CMP with $m_N = 20$ neurons and $m_C = 7$ mixture components. (\textbf A) Tuning curves of the ground truth CMP\@. (\textbf B) Stimulus-dependent mixture weights of the ground truth CMP\@, with two weight-curves highlighted in black. (\textbf C) Best of 10 descents of the negative log-likelihood by the EM (red) SGD (blue) and Hybrid (purple) algorithms, with lower- and upper-bounds given by the ground truth CMP and the 1-component model, respectively. (\textbf D) Weights of the CMP learned by the hybrid algorithm, with two weight-curves highlighted that are qualitatively similar to those in (B). (\textbf E) Stimulus-dependent, paired correlation matrices of the ground truth and hybrid-trained models. Upper-right and lower-left triangles are correlations of the ground truth and learned CMP, respectively. Bottom, middle, and top matrices are conditioned on $z=0\pi$, $z=0.33\pi$, and $z=0.67\pi$, respectively, matching axis ticks in (D).}\label{fig:synthetic-data-1}
\end{figure}

In this section we model synthetic and real data with a CMP\@. In both cases we define the context variable as an element of the half-circle such that $\mathcal Z = [0,\pi]$, which represents e.g.\ the orientation of a grating, as is common in V1 experiments. We define the natural parameter function by $\eprms_{N \mid Z}(z) = \iprms_{NZ} \cdot \V s_Z(z)$, where $\V s_Z(z) = (\cos(2z), \sin(2z))$. This choice of $\eprms_{N \mid Z}$ allows us to express the conditional firing rates of the model neurons as $\E_P[N_k \mid Z = z, C = j] = \gamma_{j,k}f_k(z)$, where $\gamma_{j,k}$ is the gain and $f_k(z) \propto e^{\rho_k \cos (x - \mu_k)}$ is a von Mises density function with preferred stimulus $\mu_k$ and precision $\rho_k$. The dependence of the model on the categorical variable is expressed solely through gain components $\V \gamma_j = {(\gamma_{j,k})}_{k=1}^{m_N}$, and where $\E_P[N_k \mid Z = z]$ is the \emph{tuning curve} of the $i$th neuron, this form ensures that each tuning curve retains a von Mises shape regardless of correlation structure.

All simulations were run on a Lenovo P51 laptop with an Intel Xeon processor, and individual simulations of the algorithms in Subsection~\ref{sec:training-algorithms} took on the order of tens of seconds to execute. All algorithms were implemented in the Haskell programming language.

\subsection{Synthetic Data}

We first consider synthetic data generated from a CMP $P_{CN \mid Z}$ with $m_N = 20$ neurons and $m_C = 7$ mixture components. The preferred stimuli $\mu_k$ of the tuning curves are drawn from a uniform distribution over $\mathcal Z$, and the precisions $\rho_k$ and gains $\gamma_{j,k}$ are log-normally distributed with a mean of 0.8 and 2, respectively. We plot the von Mises part $f_k$ of each CMP tuning curve in Figure~\ref{fig:synthetic-data-1}A. In Figure~\ref{fig:synthetic-data-1}B we plot $p_{C \mid Z}(j \mid z)$ for every $j$, as a function of $z$. This provides a low-dimensional picture of the stimulus-dependence of the correlations; the number of active components determines the dimensionality of the correlations, and their identity determines the structure of correlations.

To synthesize a dataset consistent with typical recordings in visual cortex we synthesize 62 responses to 8 stimuli tiled evenly over the half-circle, for a total of $m_S = 496$ sample points ${\{(N_i,Z_i)\}}_{i=1}^{m_S}$ from $P_{N \mid Z}$. In Figure~\ref{fig:synthetic-data-1}C we plot the descent of the resulting negative log-likelihood of the CMP parameters for each of the three algorithms proposed in Subsection~\ref{sec:training-algorithms}. Each epoch of each algorithm involves exactly one pass over the dataset, ensuring that their computation time is of a similar order. In each epoch the dataset is broken up into randomized minibatches of 50 samples, resulting in ten gradient steps per epoch for each algorithm, except for the closed-form phase of the Hybrid algorithm which processes the data as a single batch. Gradient descents are implemented with the Adam algorithm~\cite{kingma_adam_2014} with standard momentum parameters and a learning rate of 0.005, and the Adam parameters are reset every epoch. To avoid local minima we run 10 descents in parallel for each algorithm and select the best one.

In addition, we plot the true negative log-likelihood $-\sum_{i=1}^{m_S}\log p_{N \mid Z}(N_i \mid Z_i)$ in Figure~\ref{fig:synthetic-data-1}C to establish a training lower-bound. Since a CMP $Q^0_{N \mid Z}$ with $m_C + 1 = 1$ component reduces to a set of conditionally independent Poisson neurons, and fitting such a CMP is a convex optimization problem, we also plot an upper-bound in the form of $-\sum_{i=1}^{m_S}\log q_{N \mid Z}^0(N_i \mid Z_i)$ for the optimal $Q^0_{N \mid Z}$. This upper-bound tells us how much the addition of correlations to the model improves the quality of the fit. As seen in Figure~\ref{fig:synthetic-data-1}, the Hybrid algorithm converges more quickly and to a lower value than either SGD or EM\@. All algorithms overfit the data relative to the lower-bound, which is unsurprising given the small sample size. We later apply cross-validation in our analysis of real data to help avoid this.

\begin{figure}
    \includegraphics[width=0.33\textwidth]{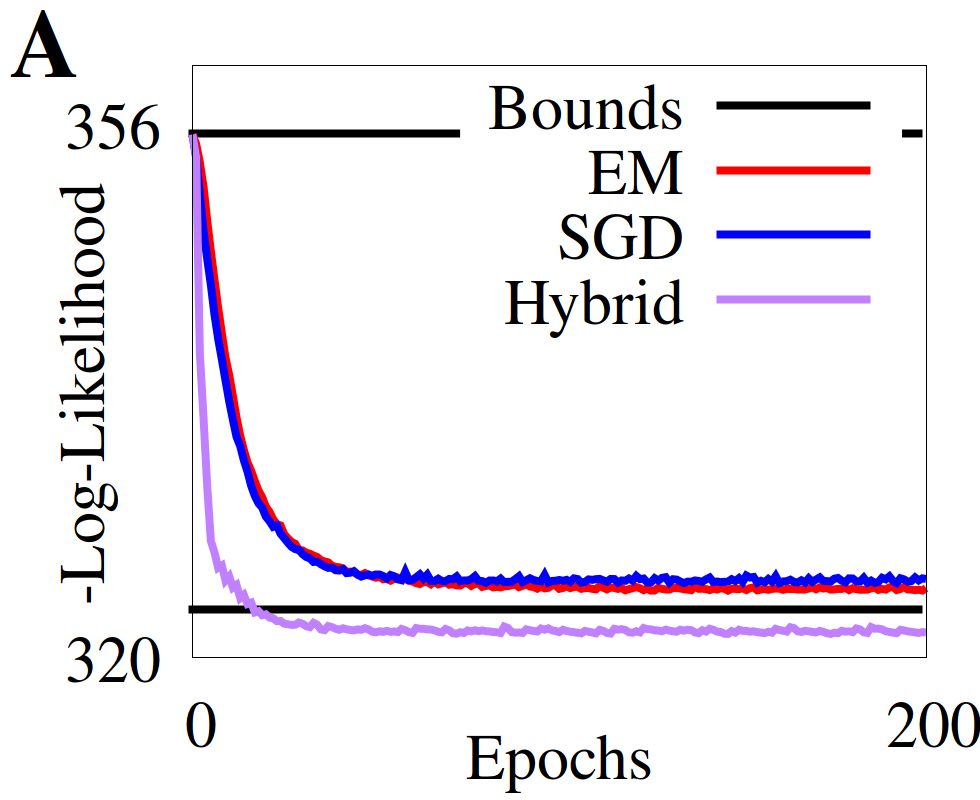}
    \includegraphics[width=0.33\textwidth]{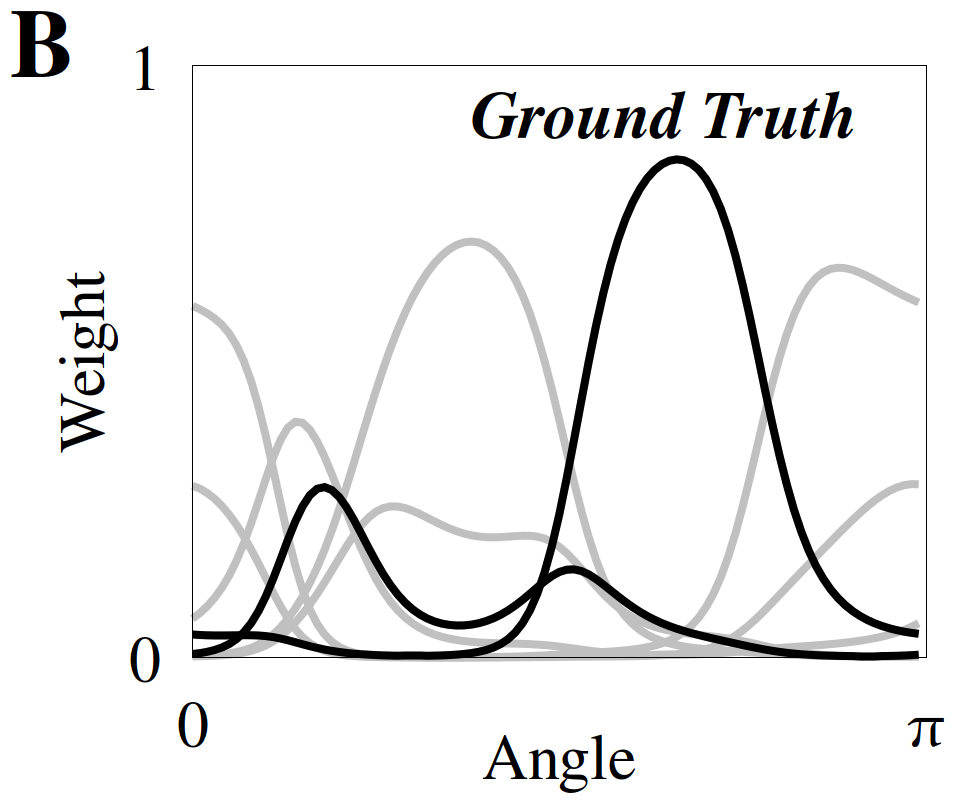}
    \includegraphics[width=0.33\textwidth]{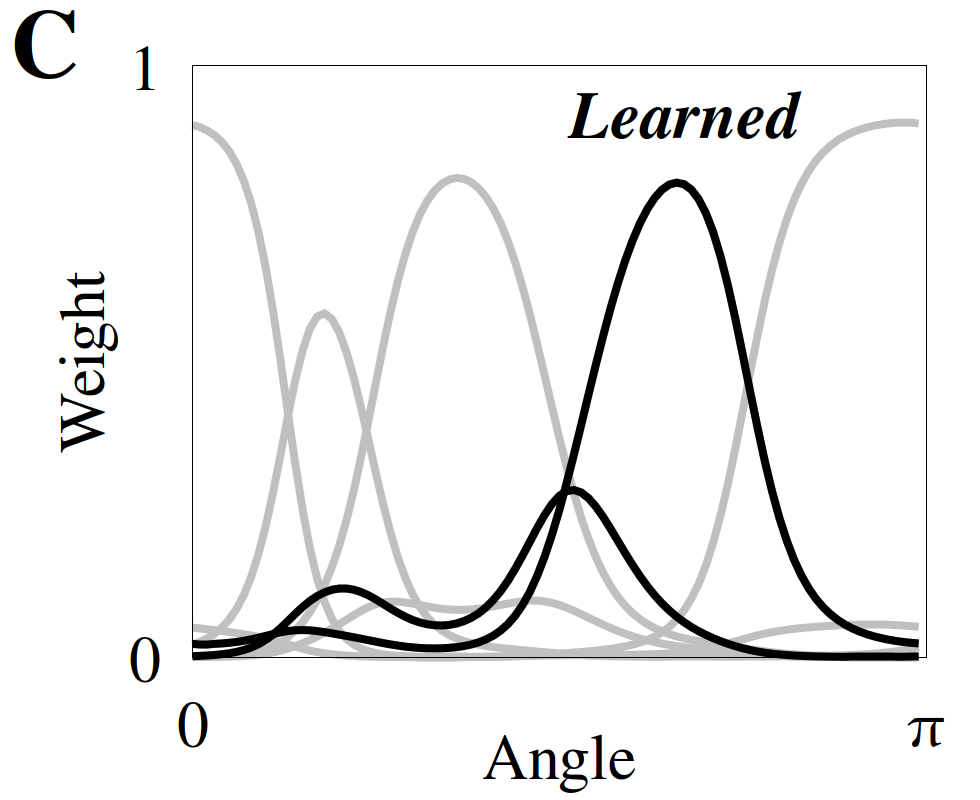}
    \caption{Training a CMP model on synthetic data generated from a ground truth CMP with $m_N = 200$ neurons. (\textbf A) Negative log-likelihood descents as per figure~\ref{fig:synthetic-data-1}C. (\textbf{B \& C}) Mixture weights of the ground truth CMP (B) with two weight-curves matched qualitatively with weight-curves from the hybrid-train CMP model (C).}\label{fig:synthetic-data-2}
\end{figure}

In Figure~\ref{fig:synthetic-data-1}D we plot $q_{C \mid Z}(j \mid z)$ for the CMP model trained with the Hybrid algorithm. We highlight two curves to emphasize that the weight-curves learned by model are similar to those of the ground truth CMP\@. In Figure~\ref{fig:synthetic-data-1}E we plot stimulus-dependent, paired correlation matrices of the ground truth and hybrid-trained models. As can be seen, the learned-model and ground truth correlations are nearly identical. Also note that when one mixture weight dominates, the correlations disappear. Finally, in Figure~\ref{fig:synthetic-data-2} we repeat the previous analysis on a CMP target and model with $m_N = 200$ neurons. As can be seen, the results are essentially identical to those in figure~\ref{fig:synthetic-data-1}. As such, at least for synthetic data, a small number of sample points (e.g. 496) are sufficient for modelling low-dimensional noise correlations in large populations of neurons.

\subsection{Response Recordings from Macaque Primary Visual Cortex}

In this subsection we repeat our analysis from the previous subsection on multi-electrode recordings of macaque primary visual cortex (V1) in response to the oriented grating stimuli (the data were originally presented in~\cite{coen-cagli_flexible_2015}). We analyze 8 datasets corresponding to 8 multi-electrode recording sessions of between 28--76 well-tuned neurons. Each dataset is composed of 80 repeated trials of 16 stimuli\footnote{Each set of 80 trials pools 4 distinct phases of the stimulus, which could inflate measured correlations. Nevertheless, recorded neurons were found to be roughly phase independent~\cite{coen-cagli_flexible_2015}.}, where the stimuli are spread evenly over the half circle.

\begin{figure}
    \includegraphics[width=0.32\textwidth]{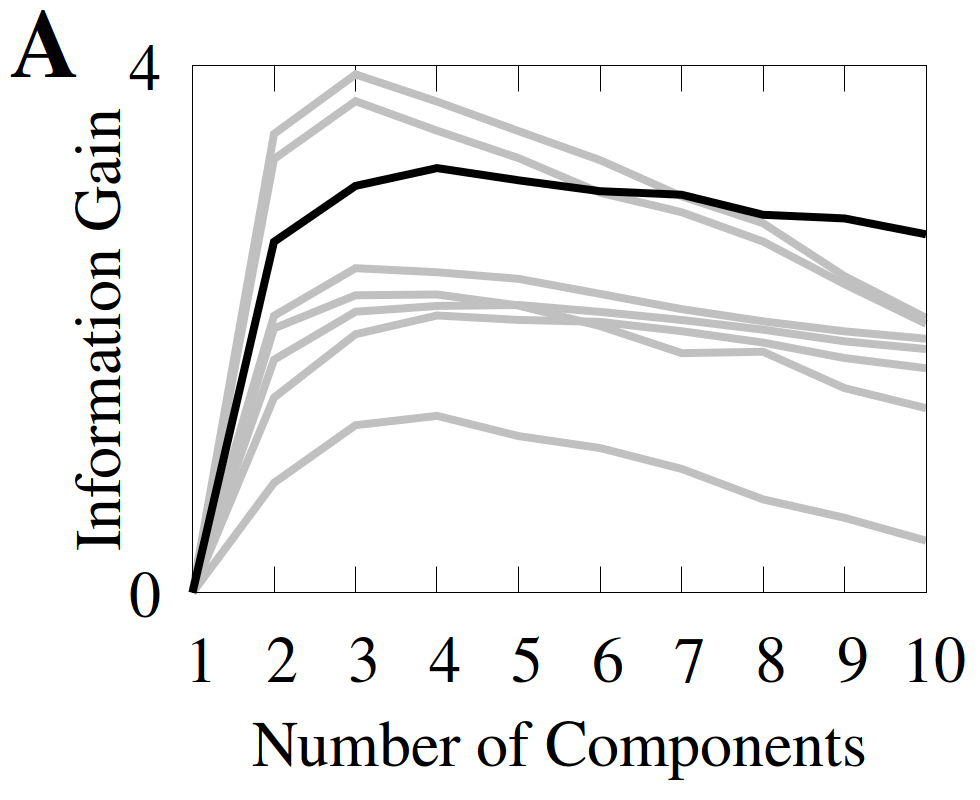}
    \includegraphics[width=0.32\textwidth]{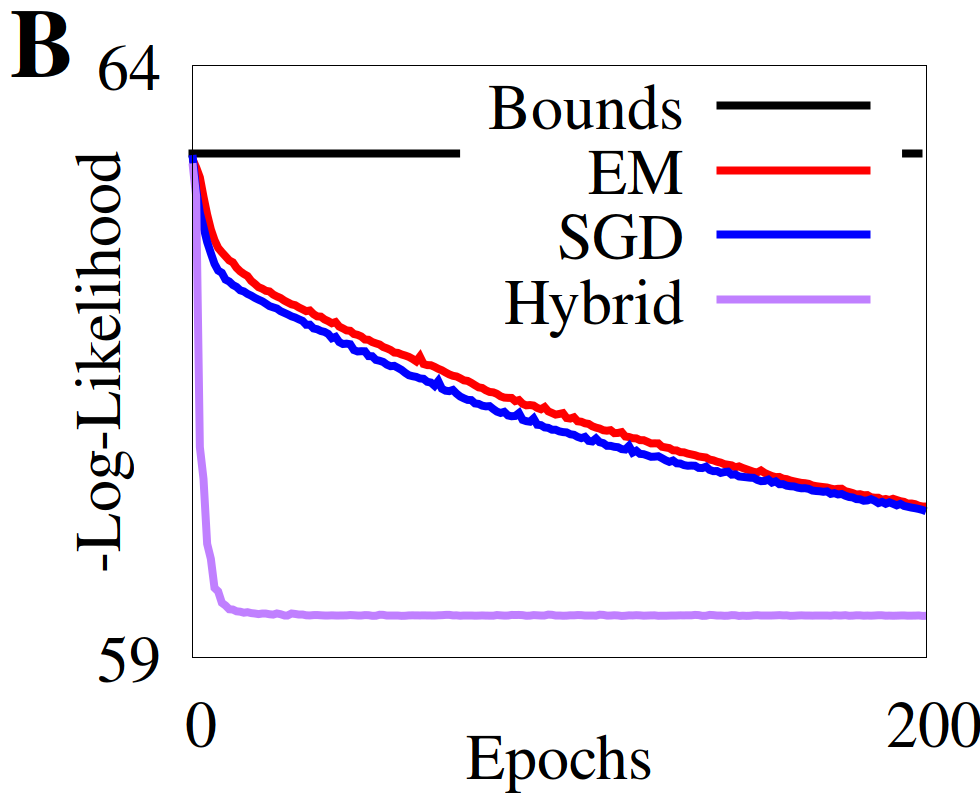}
    \includegraphics[width=0.32\textwidth]{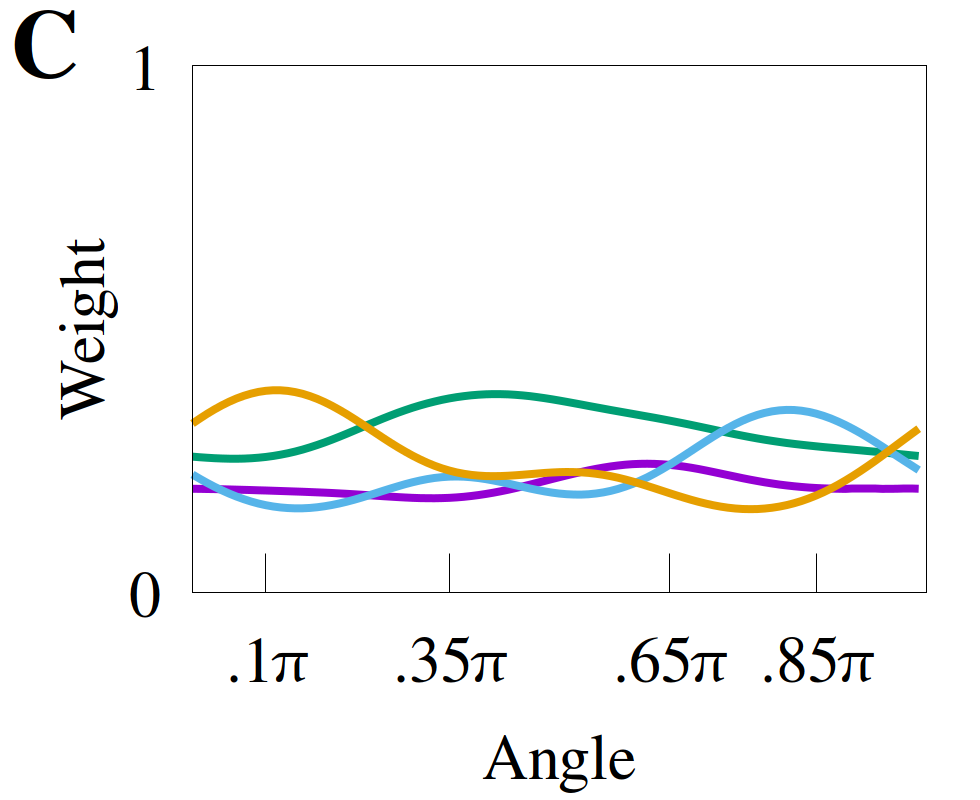}
    \begin{center}
    \includegraphics[width=0.24\textwidth]{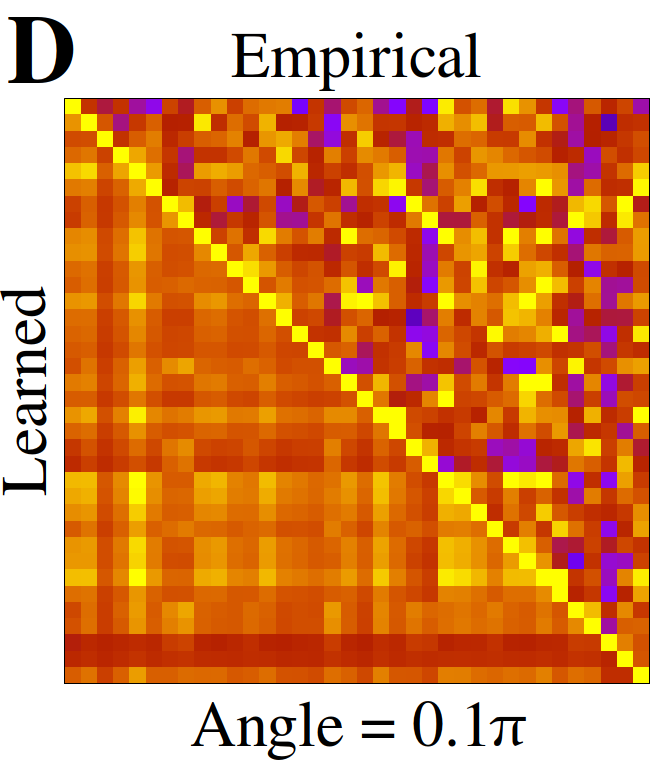}
    \includegraphics[width=0.22\textwidth]{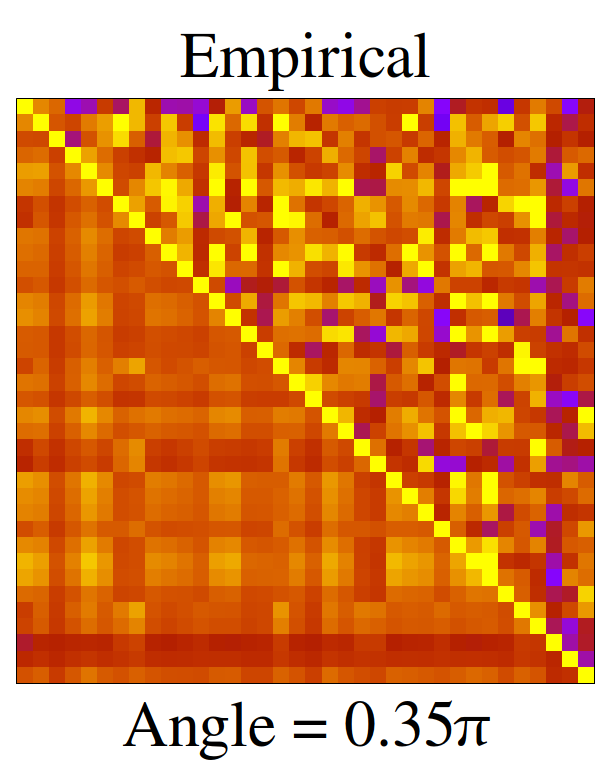}
    \includegraphics[width=0.22\textwidth]{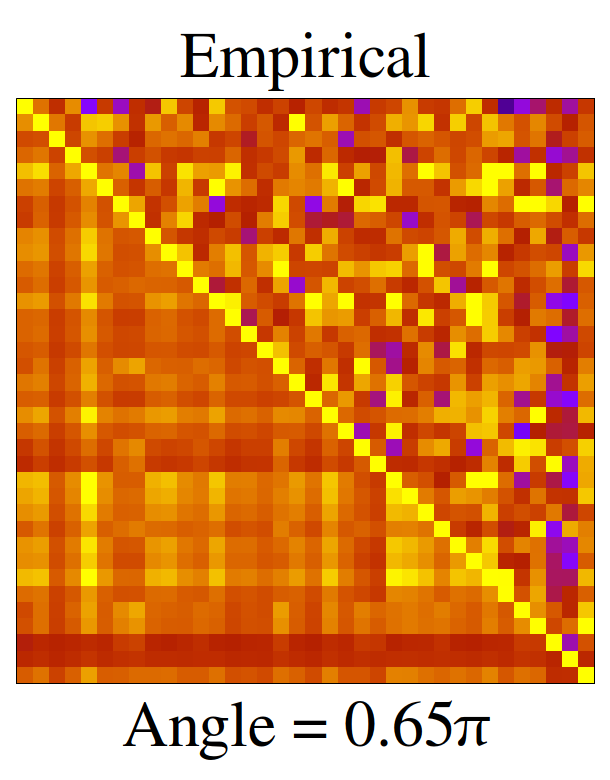}
    \includegraphics[width=0.28\textwidth]{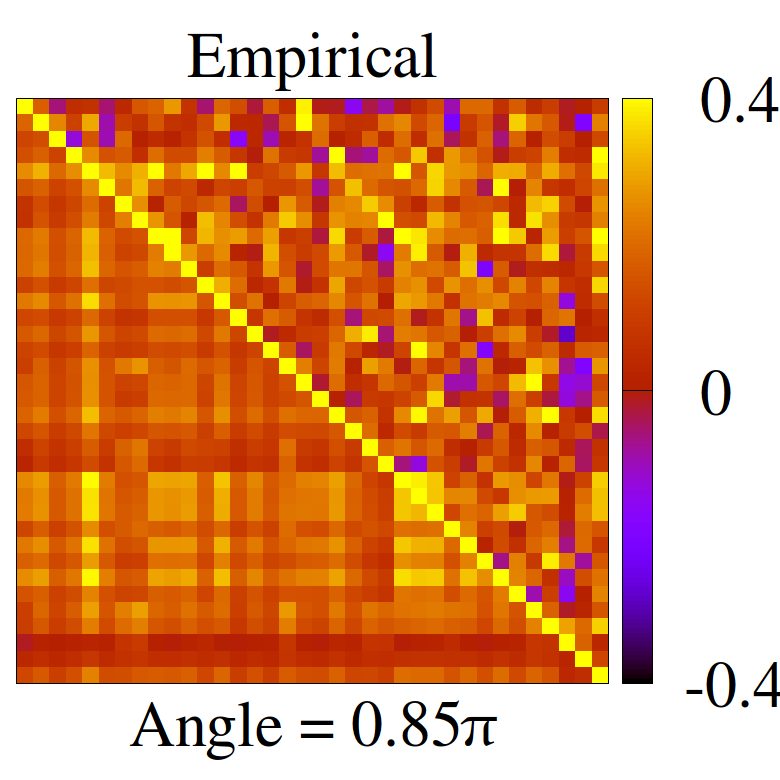}
    \end{center}
    \caption{Training a CMP model on response recordings from macaque V1. (\textbf A) 10-fold cross-validated log-likelihood of 8 datasets as a function of the number of mixture components, relative to the 1-component log-likelihood for that dataset. We highlight the curve of a selected dataset which underlies the rest of the plots in this figure. (\textbf B) Negative log-likelihood descents on the full selected dataset, as per figure~\ref{fig:synthetic-data-1}C. (\textbf C) Stimulus-dependent mixture weights of the CMP learned on the selected dataset trained with the hybrid algorithm, with tick marks indicating the conditioning angle of the correlation matrices in (D). (\textbf D) Paired correlation matrices, where the upper-right and lower-left triangles are the empirical and learned CMP correlations, respectively. Matrices are conditioned on $x=0.1\pi$, $x=0.35\pi$, $x=0.65\pi$, and $x=0.85\pi$, respectively, matching axis labels in (C).}\label{fig:coen-cagli-2015}
\end{figure}

In Figure~\ref{fig:coen-cagli-2015}A we depict the 10-fold cross-validation of the log-likelihood of each of the 8 datasets as a function of the number of components $m_C+1$ in the Hybrid-trained CMP model, and we subtract the 1-component value from the value computed for the remaining components. This figure thus quantifies the amount of information (in nats) that is gained about the true data distribution by modelling the data with the given number of components. As can be seen, between 3--5 components is optimal for the various datasets before information gain yields to overfitting. In Figure~\ref{fig:coen-cagli-2015}B we plot the results of fitting the complete selected dataset with the three proposed algorithms. The Hybrid algorithm continues to perform well and converge quickly, while it appears as though EM and SGD require significantly more computation to find good solutions.

In Figure~\ref{fig:coen-cagli-2015}C we plot $q_{C \mid Z}(j \mid z)$ for the hybrid-trained model $Q_{CN \mid Z}$, but in this case we colour the four curves to more easily distinguish them. The curves indicate that the correlations of the true data distribution are stimulus-dependent, but less so than those of our random models. Moreover, at no stimulus value does a single component dominate, and the effective dimensionality is largely stimulus-independent. Figure~\ref{fig:coen-cagli-2015}D shows that the CMP finds a subtle yet significant amount of stimulus-dependence in model correlations, which are not clearly visible in the empirical correlations.

\section{Conclusion}\label{sec:conclusion}

In this paper we introduced a novel model (CMPs) of context-dependent neural correlations, derived an expectation-maximization method for training it, and demonstrated that CMPs can capture significant correlation structure amongst neurons. We also demonstrated that EM and SGD can often perform equally well in the context of finite mixture models, which stands in contrast with common practice~\cite{mclachlan_finite_2019}. In future work we will compare the correlations learned with a CMP to the results of alternative dimensionality-reduction techniques for context-dependent neural correlations~\cite{granot-atedgi_stimulus-dependent_2013,ecker_state_2014,cowley_stimulus-driven_2016}. At the same time, CMPs are uniquely suited for studying rate-based neural coding. For example, information-limiting correlations~\cite{moreno-bote_information-limiting_2014} can be directly incorporated into a CMP by defining the gain components as identical but shifted with respect to neuron index.

In our applications in Section~\ref{sec:applications}, the stimulus-dependent parameters had a simple generalized linear form. Theoretically, however, these parameters could be modelled by a deep neural network, thereby allowing the output of the deep network to exhibit correlations. When the output of the network has a complex structure, such as when modelling the stimulus-dependent neural responses of mid-level brain regions to naturalistic images~\cite{yamins_using_2016}, incorporating correlations could significantly improve model predictions. Moreover, because CMPs model noise correlations in a manner that is compatible with theories of population coding, such a combined model can extend our understanding of neural computation beyond low-level sensory areas and low-dimensional variables.

\bibliographystyle{unsrt}
\small{\bibliography{library}}

\end{document}